\documentclass[11pt]{article}

\usepackage[final]{acl}

\usepackage{times}
\usepackage{latexsym}

\usepackage[T1]{fontenc}
\usepackage[utf8]{inputenc}

\usepackage{microtype}

\usepackage{inconsolata}

\usepackage{amsmath}
\usepackage{amssymb}

\usepackage{graphicx}

\newcommand{\eat}[1]{}

\usepackage{amsfonts}
\usepackage{colortbl}
\usepackage{xspace}
\usepackage{comment}
\usepackage{booktabs}
\usepackage{multirow}
\usepackage{framed}
\usepackage{subcaption}

\usepackage{circledsteps}
\usepackage{fvextra}

\definecolor{green}{RGB}{0,128,0}

\definecolor{yellow}{RGB}{255,200,18}

\newcommand{\add}[1]{\textcolor{black}{{#1}}}

\newcommand{\stab}{\vspace{1.2ex}\noindent}

\newcommand{\bi}{\begin{itemize}}
\newcommand{\ei}{\end{itemize}}

\newcommand{\be}{\begin{enumerate}}
\newcommand{\ee}{\end{enumerate}}
\newcommand{\beqn}{\begin{eqnarray*}}
\newcommand{\eeqn}{\end{eqnarray*}}

\newcommand{\stitle}[1]{\stab\noindent{\bf #1}}

\newcommand{\ie}{{\em i.e.,}\xspace}

\usepackage{listings}

\usepackage{pifont}
\usepackage{tikz}
\usepackage{enumitem}

\usepackage{tcolorbox}
\tcbuselibrary{skins, breakable, theorems}

\usepackage{placeins}
\usepackage{float}

\newcommand{\method}{{\sc \textbf{EvoQuant}}\xspace}

\title{EVOQUANT: Self-Evolving Verifier-Guided Strategy Optimization\\for Robust Quantitative Trading}

\author{
  \textbf{Jie Mao}\textsuperscript{1,$\ast$},
  \textbf{Changlun Li}\textsuperscript{1,2,$\ast$},
  \textbf{Xiang Li}\textsuperscript{1},
  \textbf{Qiqi Duan}\textsuperscript{2},
  \textbf{Jinhui Yuan}\textsuperscript{1},
  \textbf{Xiang Liu}\textsuperscript{1}, \\
  \textbf{Yuyu Luo}\textsuperscript{1,2},
  \textbf{Jing Tang}\textsuperscript{1},
  \textbf{Nan Tang}\textsuperscript{1,2}
\\[4pt]
  \textsuperscript{1}The Hong Kong University of Science and Technology (Guangzhou) \\
  \textsuperscript{2}Paradoox AI Research \quad
\\[2pt]
  \small{$\ast$ Equal contribution. \quad
        Corresponding author: \texttt{research@paradoox.ai}}
}

\begin{document}
\maketitle


\begin{abstract}
Quantitative strategy optimization remains largely manual, requiring domain experts to identify weak signals, tune risk-control rules, and repeatedly validate iterative revisions. Large language models can accelerate this process, but directly relying on them to rewrite trading strategies often introduces hallucinated edits, strategy drift, and backtest overfitting.
We propose \method{}, a self-{\bf E}volving {\bf V}erifier-guided framework for strategy {\bf O}ptimization in {\bf Quant}itative trading. Our method utilize LLMs to deeply diagnoses performance bottlenecks, generates semantically controlled candidate edits, selects the best strategy through a multi-stage verification pipeline, and distills optimization experience into reusable knowledge for continual self-improvement. We evaluate our method using seven representative strategies: four from the A-share market and three from the Crypto market. Experimental results show that our method significantly improves the Sharpe ratio across all tested strategies: the average test Sharpe increases from \(-0.298\) to \(0.538\), and the best-performing strategy achieves a \(199\%\) relative improvement. Ablation studies and stress tests under stricter conditions further validate the effectiveness and robustness of the framework.
Overall, this work transforms quantitative strategy optimization from costly manual trial-and-error into an automated and verifiable iterative paradigm, offering a new path for applying large language models to financial strategy research.
\end{abstract}


\section{Introduction}
\begin{figure*}[t]
\centering
\includegraphics[width=\textwidth]{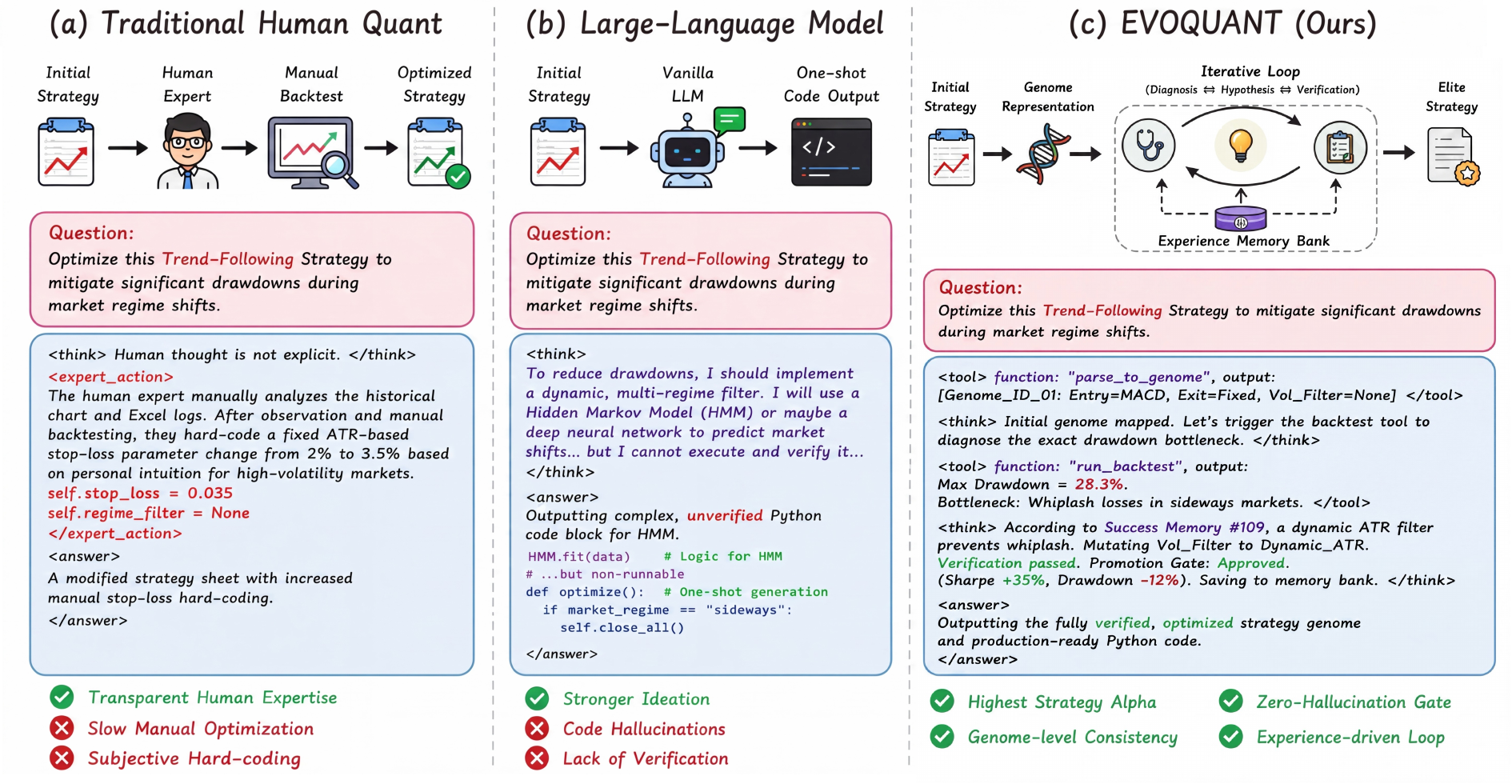}
\caption{Motivation. Compared with manual human optimization and vanilla LLM rewriting, \method{} provides a verifier-guided closed loop for controlled and reliable strategy improvement.}
\label{fig:method_flow}
\end{figure*}

Quantitative strategy development is fundamentally an iterative scientific process: researchers propose a testable market hypothesis, implement a trading rule, validate its performance through rigorous backtesting, diagnose failure modes, revise the hypothesis, and repeat this cycle continuously to adapt to non-stationary market regimes~\cite{noormohammadzadehmaleki2026artificial,zhou2025deltalag,li2025time,li2025finkario}. 
%
As summarized in Figure~\ref{fig:method_flow}(a), traditional manual optimization of quantitative strategies is constrained by scarce domain expertise, repetitive labor-intensive analysis, low iteration efficiency, and performance bottlenecks that are difficult to overcome~\cite{wang2023alphagpt,yu2025finmem}.

As shown in Figure~\ref{fig:method_flow}(b), Modern large language model (LLM) agents promise to dramatically accelerate this iterative loop by automating code generation, hypothesis testing, and strategy refinement~\cite{luo2025large,tang2025agentbuilder,dong2025survey}. Nevertheless, the direct application of general-purpose large language models (LLMs) to quantitative finance introduces new critical challenges. Specifically, these models are prone to generating hallucinatory trading rules, producing unverifiable and unexecutable strategy modifications, and failing to provide transparent causal explanations for their proposed changes~\cite{wu2025natural,li2024dawn}.  A system that generates many candidates can look impressive by exploiting noisy validation windows, low trade counts, or unreported degrees of freedom~\cite{white2000reality,bailey2014probability,harvey2016and}. Therefore, \textbf{the central problem is not simply to generate strategies, but to improve strategies under a protocol that records, verifies, and rejects unsafe edits.}
\add{Throughout this process, the LLM acts only as a bounded edit proposer over the typed strategy genome; all data-dependent accept or reject decisions are made by deterministic verification and promotion modules, and the model is never trained on market data.}

Existing LLM-driven quantitative finance studies focus on autonomous trading strategy construction and alpha factor mining~\cite{kou2024automate,tang2025alphaagent,wang2026factorminer,li2026r,tian2025agent}. In contrast, we study a more practically impactful yet understudied problem: the systematic improvement of existing deployed quantitative strategies. Figure~\ref{fig:method_flow}(c) illustrates the workflow and advantages of our proposed optimization framework. Our framework takes a user-provided quantitative strategy as input. 
Unlike unconstrained generative search, we formulate automated strategy optimization as a closed-loop scientific process. In this process, the system explicitly represents the underlying hypothesis, diagnoses current failure modes, guides the LLM to generate controlled optimization hypotheses step by step, and adopts a modification only after it survives a rigorous multi-stage verification pipeline.


The proposed method implements this design through four technical principles.
(i) It converts raw strategy logic into a typed strategy genome. This genome serves as an editable intermediate representation that supports interventions ranging from parameter tuning and local structural repair to logic rewriting. 
(ii) It builds an LLM-based strategy risk-diagnosis agent. By integrating backtest evidence across multiple time periods, the agent analyzes failure modes from the perspectives of trading behavior, risk exposure, and market compatibility, thereby identifying the core weaknesses of the current strategy.
(iii) Candidate generation is conditioned on the diagnostic results and organized into a hierarchy of increasingly expressive edits. Guided by the LLM, the system progresses from conservative local tuning to structural reconstruction, strategy-family migration, and hypothesis rewriting. Each candidate is emitted as an explicit edit plan over the strategy genome, enabling drift constraints during optimization and post-hoc provenance auditing.
(iv) An out-of-sample-aware verifier and an adaptive promotion engine jointly screen candidates through hard admission rules and robustness penalties. A candidate replaces the current incumbent only after passing all verification stages. 
In addition, a built-in knowledge-distillation module continuously distills effective experience from the optimization process into a reusable strategy knowledge base, enabling continual self-improvement of the system.

We evaluate whether the proposed iterative process improves quantitative strategies across both equity and cryptocurrency markets while mitigating single-asset artifacts and optimization overfitting.
Across seven representative strategies, the system consistently improves test performance, raising the average Sharpe ratio from $-0.298$ to $0.538$, with the best-performing strategy achieving a $199\%$ relative improvement.
Ablation and stress-test results further show that diagnosis, LLM-based candidate generation, adaptive promotion, and reusable memory each contribute to robust optimization gains.

Our contributions are:
\begin{itemize}[leftmargin=*]
  \item \textbf{Evidence-constrained strategy evolution.} We formulate automated quantitative strategy improvement as evidence-constrained program evolution, where user-provided strategies are represented as structured trading genomes that support localized, executable, and provenance-preserving edits.
  \item \textbf{LLM-enabled verifier-guided optimization protocol.} We develop a verifier-guided optimization framework that uses LLMs for failure-mode diagnosis and hierarchical candidate synthesis, while integrating out-of-sample-aware verification, adaptive promotion, and memory-based experience reuse into a unified auditable loop.
  \item \textbf{Long-horizon cross-market evaluation.} We provide a long-horizon empirical evaluation across A-share and Bitcoin strategies, including before--after performance comparisons, component ablations, transaction-cost stress tests, and walk-forward validation.

\end{itemize}
\section{Related Work}

\stitle{Backtest overfitting and data snooping.}
Financial machine learning has long recognized that repeated strategy search can inflate apparent performance through data snooping and backtest overfitting~\cite{white2000reality,bailey2014probability}. Practical treatments emphasize leakage-aware validation~\cite{lopezdeprado2018advances}, and recent theory quantifies in-sample to out-of-sample Sharpe decay~\cite{jacquier2025insample}. Our work addresses this concern by logging rejected candidates, exposing low-trade failures, tracking validation--test behavior, and reporting non-degradation rather than only the best discovered strategy.

\stitle{LLM agents and iterative refinement.}
LLM agents can reason, act, and revise outputs through feedback loops~\cite{yao2022react,shinn2023reflexion,madaan2023selfrefine}, and recent finance agents extend this pattern to market analysis, factor mining, and tool-augmented trading~\cite{zhang2024multimodal,li2024fama,zhang2025llmalphasurvey}. Newer benchmarks and critiques highlight instability and weak deployment evidence in LLM trading agents~\cite{zhang2026alphaforgebench,ye2026alphaillusion}. \method{} differs by grounding each strategy edit in chronological backtests, trade-coverage checks, cost assumptions, and risk-sensitive promotion.

\stitle{Programmatic strategy search.}
Programmatic strategy search has moved from evolutionary and symbolic alpha discovery~\cite{kakushadze2016formulaic,cu2021alphaevolve} toward automated formulaic-alpha mining with reinforcement learning, LLM agents, structure-aware search, and dedicated evaluation protocols~\cite{shi2024alphaforge,zhao2024quantfactor,tang2025alphaagent,ding2025alphaeval,chen2025alphasage,yang2026alphacfg,wang2026factorminer}. These approaches scale candidate generation and assessment, but broad search spaces can blur whether a gain reflects principled refinement of an initial market hypothesis or selection among many degrees of freedom. \method{} instead treats optimization as constrained program evolution from a user-provided strategy, with typed genome edits, diagnosis-conditioned generation, out-of-sample-aware promotion gates, and an explicit audit trail.
\section{Methodology}
\begin{figure*}[t]
\centering
\includegraphics[width=\textwidth]{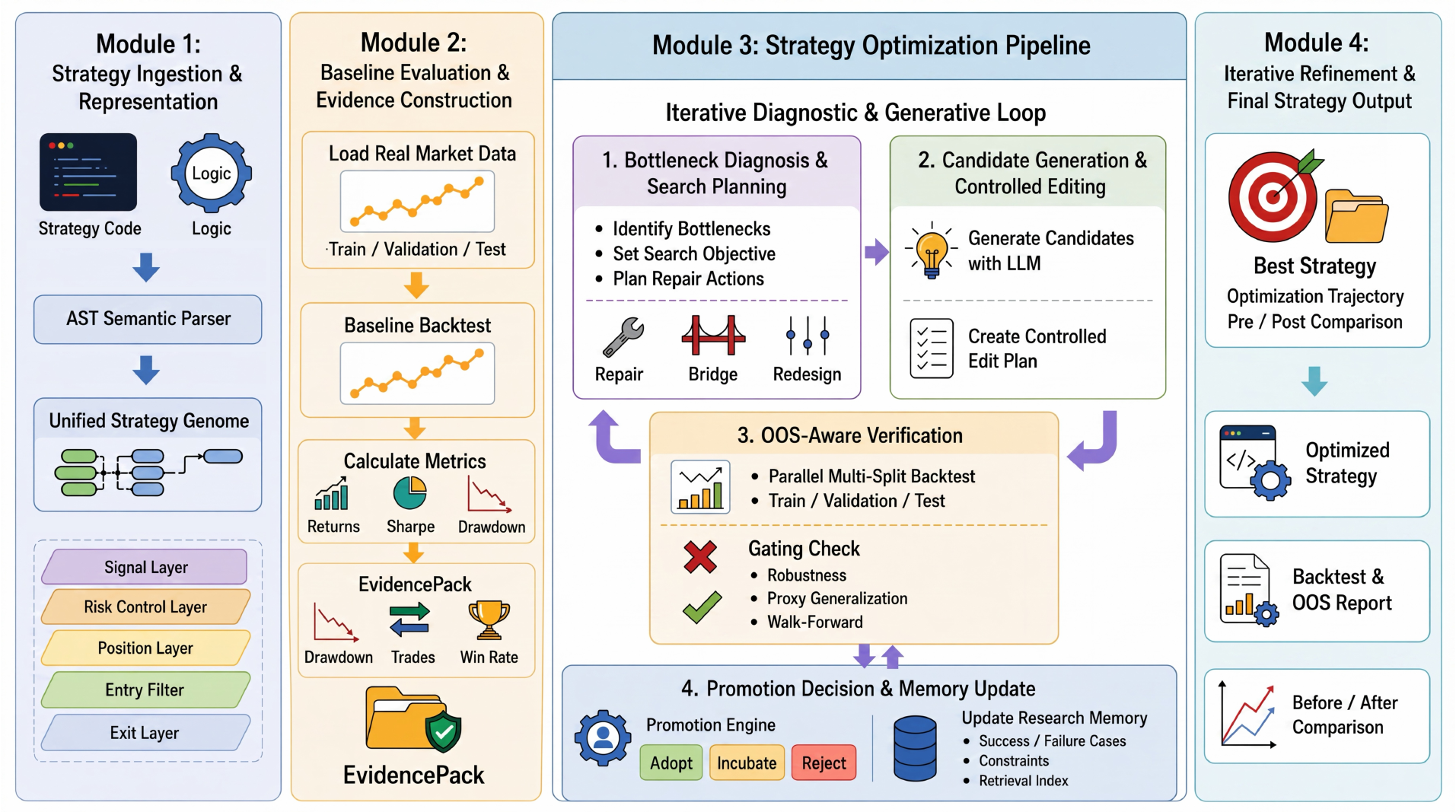}
\caption{\method{} workflow organized into four modules. Module 1 normalizes raw strategy code or logic into a typed strategy genome. Module 2 constructs chronological baseline evidence. Module 3 performs the iterative diagnostic--generative loop, including bottleneck diagnosis, controlled candidate editing, verification, and promotion with memory update. Module 4 outputs the best verified strategy, optimization trajectory, and report.}
\label{fig:method_flow1}
\end{figure*}

\method{} casts quantitative strategy optimization as \emph{verifier-guided program evolution}. As shown in Figure~\ref{fig:method_flow1}, the system contains four modules: strategy ingestion and representation, baseline evidence construction, a diagnostic--generative optimization loop, and final strategy refinement/output. The core design of our framework emphasizes practical execution over pure generation. The large language model proposes strategy revisions within controllable scopes, and components including the backtester and verifier jointly determine whether such revisions are admissible.

At round \(t\), the incumbent genome \(g_t\) is evaluated into an evidence pack \(E_t\). The optimization loop diagnoses the current bottleneck, generates candidates \(\mathcal{C}_t\), and promotes only a verified candidate:
\[
  g_{t+1} =
  \begin{cases}
    c^\star_t, & \text{if } c^\star_t \text{ is accepted by } \Pi,\\
    g_t, & \text{otherwise}.
  \end{cases}
\]
where \(c^\star_t\) is the best verified candidate selected from \(\mathcal{C}_t\), and \(\Pi\) denotes the promotion rule later specified in Module~3. 
The loop adopts a conservative design: failed candidates are stored with rejection reasons and later used to avoid repeating unproductive edits.

\subsection{Module 1: Strategy Ingestion and Representation}
\method{} first normalizes the user strategy into a typed intermediate representation called a \emph{strategy genome}. This module performs four steps. First, it parses raw code or rule text into an abstract syntax tree. Second, it tags trading semantics such as signals, entry conditions, risk rules, position sizing, and exits. Third, it factors the strategy into editable layers. Finally, it recompiles the genome to ensure that the representation remains executable. Formally, the parser \(\Phi\) maps a strategy \(s\) to
\[
  \Phi(s)=g=
  \big(g^{\mathrm{sig}},g^{\mathrm{risk}},g^{\mathrm{pos}},
  g^{\mathrm{entry}},g^{\mathrm{exit}},\xi\big),
\]
where \(\xi\) stores provenance, strategy family, parent identifiers, and mutation history. The output of this module is not merely a parsed program, but an editable research object. Later edits must specify the target layer, satisfy non-anticipativity and exposure constraints, and compile back to executable strategy semantics \(g\).

\subsection{Module 2: Baseline Evaluation and Evidence Construction}
\method{} next measures the original strategy before attempting to improve it. The evaluator loads real market data, constructs chronological splits, and runs the incumbent genome under the same execution assumptions used for all future candidates. It records both scalar metrics and behavioral profiles, producing

\[
\begin{aligned}
E(g)=&
\Big\{
M_X,\; P_X^{\mathrm{trade}},\; P_X^{\mathrm{regime}}
\Big\}_{X\in\{\mathrm{train},\mathrm{val},\mathrm{test}\}} \\
&\cup
\{\delta_{\mathrm{oos}},\; C_{\mathrm{conc}},\; H_{\mathrm{risk}}\}.
\end{aligned}
\]
Here, \(M_X\) comprises performance metrics such as return, Sharpe ratio, maximum drawdown, trade count, win rate, and related statistics. \(P_X^{\mathrm{trade}}\) characterizes trading behaviors including trade sparsity, holding period distributions, and return concentration, while \(P_X^{\mathrm{regime}}\) quantifies the strategy's sensitivity to different market regimes. The metric \(\delta_{\mathrm{oos}}\) reflects the degradation from validation to out-of-sample (OOS) performance, and \(H_{\mathrm{risk}}\) documents abnormal risk phenomena, such as excessive stop-loss triggering or truncated upside returns. This module provides comprehensive evidence for the downstream diagnostic loop, enabling the system to assess not only whether a strategy underperforms, but also to elucidate the specific failure modes underlying any observed weaknesses.

\subsection{Module 3: Strategy Optimization Pipeline}
\stitle{Bottleneck Diagnosis and Search Planning.}
It starts by turning Evidence pack statistics into a bottleneck diagnosis and a concrete search plan.
\[
  \begin{aligned}
  z_t &= B(E_t,\mathcal{M}_t),\\
  p_t &= (o_t,\ell_t,\mathcal{O}_t,\mathcal{F}_t,K_t).
  \end{aligned}
\]
The diagnosis is implemented by large language models through rule-based evidence verification and in-depth analysis. It identifies failure modes such as risk-rule alpha truncation, regime mismatch, validation decay and so on. The search plan then specifies the round objective \(o_t\), search level \(\ell_t\), allowed operators \(\mathcal{O}_t\), forbidden shortcuts \(\mathcal{F}_t\), and candidate budget \(K_t\).

The diagnosis constrains the generator's action space. For example, If the bottleneck is validation--OOS decay, the plan favors simplification, robustness repair, and parameter smoothing rather than adding more regime-specific conditions. The search level progresses from \emph{repair} to \emph{bridge}, \emph{redesign}, and finally \emph{family migration}. This keeps the adaptivity of AutoML and LLM-as-optimizer loops~\cite{hutter2011sequential,akiba2019optuna,yang2023opro,guo2023evoprompt}, but makes each search step accountable to a financial failure mode.

\stitle{Candidate Generation and Controlled Editing.}
Given the search plan, \method{} builds a candidate-generation prompt from four artifacts: the current genome, a compressed EvidencePack, the bottleneck diagnosis, and retrieved memory cases. The LLM must return a controlled edit plan with fields for target layer, operation, patch content, mechanism rationale, expected metric movement, and safety constraints. A candidate is accepted for testing only if it satisfies the typed edit contract:
\[
  \begin{aligned}
  c_{t,i} &= \Delta_{t,i}(g_t),\\
  \Delta_{t,i} &\in \mathcal{T}(p_t,\mathcal{M}_t),\\
  d_G(g_t,c_{t,i}) &\le \epsilon_{\ell_t},\quad
  \mathcal{I}(c_{t,i})=1 .
  \end{aligned}
\]
Here \(\Delta_{t,i}\) is an edit operator, \(d_G\) measures semantic drift from the incumbent, and \(\mathcal{I}\) checks executable and financial invariants. In practice, parameter repair adjusts lookback windows, thresholds, stop widths, and position caps; bridge edits modify entry filters, exit rules, or signal combinations; redesign rewrites the strategy graph around a coherent market mechanism. Family migration, the highest-tier modification, shifts to a nearby strategy family after repeated local failures. Before evaluation, candidates are compiled, deduplicated by genome hash, and diversified so that one round does not spend all budget on near-identical edits.

\stitle{OOS-Aware Verification, Promotion, and Memory.}
The verifier evaluates candidates in parallel under the same data split and execution assumptions as the incumbent. It first applies hard gates that are intentionally simple and difficult to game:
\[
  \begin{aligned}
  G(c)=1 \;\Longleftrightarrow\;&
  N_{\mathrm{val}}(c)\ge n_{\min},\;
  N_{\mathrm{oos}}(c)\ge n_{\min},\\
  &\mathrm{MDD}_{\mathrm{oos}}(c)\le \bar d,\\
  &d_G(g_t,c)\le\epsilon_{\ell_t},\;
  \mathcal{I}(c)=1 .
  \end{aligned}
\]
These gates remove non-executable, zero-trade, overly drifting, and high-drawdown candidates before any soft scoring. Gated candidates are then ranked with a compact score:
\[
  \begin{aligned}
  Q(c)=\;&
  w_v\Delta S_{\mathrm{val}}
  +w_o\Delta S_{\mathrm{oos}}
  +w_r\Delta R_{\mathrm{oos}}\\
  &-\lambda_d P_{\mathrm{dd}}
  -\lambda_g P_{\mathrm{gap}}
  -\lambda_s P_{\mathrm{stress}} .
  \end{aligned}
\]

Here \(S\) denotes Sharpe, \(R\) denotes return, \(P\) denotes the penalty and all deltas are candidate-minus-incumbent. The penalties capture OOS (out-of-sample) drawdown expansion, validation--OOS discrepancy, and lightweight stress probes such as fee/slippage perturbation, one-bar delay, and parameter jitter. The score deliberately separates validation improvement from out-of-sample audit behavior, so a fluent rewrite that wins only on validation is not automatically promotable.

The promotion engine assigns each candidate to \textsc{adopt}, \textsc{incubate}, or \textsc{reject}. \textsc{Adopt} replaces the incumbent only when the candidate passes hard gates and exceeds an adaptive score threshold. \textsc{Incubate} stores candidates that show a plausible mechanism but insufficient evidence, while \textsc{reject} records failure reasons such as zero trades, OOS collapse, excessive drift, or abnormal drawdown. After the decision, Memory module stores the diagnosis, search plan, edit summary, candidate evidence, decision, and failure reason. This memory is first distilled and persisted in the strategy repository, serving dual purposes in subsequent iterations: retrieving proven optimization paradigms and blacklisting recurrent failure mechanisms, thereby enabling continuous self-improvement of the system.

\subsection{Module 4: Iterative Refinement and Final Strategy Output}
The final module manages convergence and output. It tracks the best verified genome observed across the trajectory,
\[
  g^\star_t=\arg\max_{g\in\mathcal{H}_t} Q_{\mathrm{audit}}(g),
\]
where \(\mathcal{H}_t\) contains adopted and incubated candidates observed so far. If local repairs fail repeatedly, the controller escalates to bridge edits, redesign, and finally family migration. The loop stops when the round budget is exhausted, the audit score converges, or no admissible search level passes verification. The final output includes the optimized strategy code \({g^\star}\), a pre/post backtest report, and the full optimization trajectory with diagnoses, edit plans, promotion decisions, and memory records.

\section{Experimental Protocol}

\stitle{Data and LLM Backend.}
All experimental data are obtained from the AKShare database\footnote{\url{https://akshare.akfamily.xyz/}}. We use daily historical data for both the A-share market and Bitcoin from 1/1/2020 to 31/12/2025. For all components that require large language model inference, we use the DeepSeek-R1 model~\cite{guo2025deepseek}. Each optimization run lasts 20 iterations.

\stitle{A-share Main Task.}
We evaluate four representative A-share quantitative strategies: an EMA-RSI trend-continuation strategy, a Bollinger Band-RSI mean-reversion strategy, a volume-breakout strategy, and a low-volatility quality-stock strategy. We randomly sample 30 stocks as the experimental universe and independently optimize each strategy--stock pair. 
For each strategy family, the final performance is computed as an equal-weighted portfolio over the 30 optimized stock-level strategies. All core empirical conclusions for the A-share market are based on these portfolio-level backtest results.

\stitle{Bitcoin Main Task.}
We further evaluate three Bitcoin trading strategies: a MACD-RSI-Bollinger Band strategy, an ATR trend-breakout strategy, and an oversold-reversal strategy. 
This experiment examines whether the proposed optimization framework can adapt to an alternative asset class with substantially different volatility and trend characteristics, while still improving strategy performance.

\stitle{Ablation Studies.}
To isolate where the optimization gains come from, we organize the ablation study into two groups. The first group, \emph{setting variations}, changes the search or selection configuration while keeping the overall pipeline intact: \emph{Simple threshold} replaces adaptive promotion with a fixed threshold rule, \emph{Parameter only} restricts candidate generation to parameter-level tuning, \emph{Random mutation only} replaces diagnosis-conditioned generation with random edits, \add{and \emph{Bayesian Opt parameter-only} tunes numeric parameters with Bayesian optimization under the same verifier protocol but cannot rewrite strategy logic}. The second group, \emph{pipeline variations}, removes major functional modules from the full system: \emph{w/o promotion engine} disables adaptive candidate adoption, \emph{w/o rule diagnosis} removes rule-based failure-mode diagnosis, and \emph{w/o memory} disables both retrieval from and writing to historical optimization experience. All ablations are conducted on the A-share volume-breakout task (See Table~\ref{tab:ablation}).

\stitle{Robustness and Walk-Forward Tests.}
We perform robustness and walk-forward tests on the A-share volume-breakout strategy. For robustness, we double the transaction cost and slippage assumptions, and change the ``train/validation/test'' split points. For walk-forward tests, we use two distinct rolling windows (2020-2023, 2021-latest) compared to the original (2020-2025).

\section{Results}




\begin{table*}[!t]
\centering
\small
\resizebox{\textwidth}{!}{\begin{tabular}{llrrrrrrrrr}
\toprule
\multirow{2}{*}{Market} 
& \multirow{2}{*}{Strategy family}
& \multicolumn{3}{c}{Sharpe $\uparrow$}
& \multicolumn{3}{c}{Return (\%) $\uparrow$}
& \multicolumn{3}{c}{MDD (\%)$\downarrow$} \\
\cmidrule(lr){3-5}
\cmidrule(lr){6-8}
\cmidrule(lr){9-11}
& 
& Base & Opt. & $\Delta$
& Base & Opt. & $\Delta$
& Base & Opt. & $\Delta$ \\
\midrule
\multirow{4}{*}{A-share}
& Bollinger Reversion 
& 0.351 & 0.966 & 0.615
& 4.89 & 9.02 & 4.13
& 5.01 & 6.71 & 1.70 \\

& EMA-RSI Trend 
& -0.465 & 0.367 & 0.832
& 1.64 & 16.83 & 15.19
& 14.36 & 10.41 & -3.95 \\

& Low-Vol Quality 
& -0.500 & 0.386 & 0.885
& 1.58 & 17.86 & 16.28
& 9.30 & 8.26 & -1.04 \\

& Volume Breakout 
& -0.493 & 0.489 & 0.982
& -2.21 & 15.74 & 17.95
& 4.81 & 6.62 & 1.81 \\
\midrule
\multirow{3}{*}{Crypto}
& MACD-RSI Bollinger 
& -1.118 & 0.265 & 1.383
& -13.27 & 5.42 & 18.69
& 18.22 & 8.54 & -9.68 \\

& ATR Trend Breakout 
& -1.536 & -1.050 & 0.486
& -14.48 & -9.67 & 4.81
& 14.06 & 12.01 & -2.05 \\

& Oversold Reversal 
& -0.838 & 0.679 & 1.516
& -8.37 & 11.56 & 19.93
& 16.53 & 12.17 & -4.36 \\
\bottomrule
\end{tabular}}
\caption{Main results across A-share and cryptocurrency strategy families. $\Delta$ is computed as Opt.(optimized) minus Base (original). Arrows indicate the preferred direction for each metric. Positive $\Delta$ is desirable for Sharpe and Return, while negative $\Delta$ is desirable for MDD.}
\label{tab:main_astock_btc}
\end{table*}


\begin{figure*}[!t]
\centering
\includegraphics[width=\textwidth]{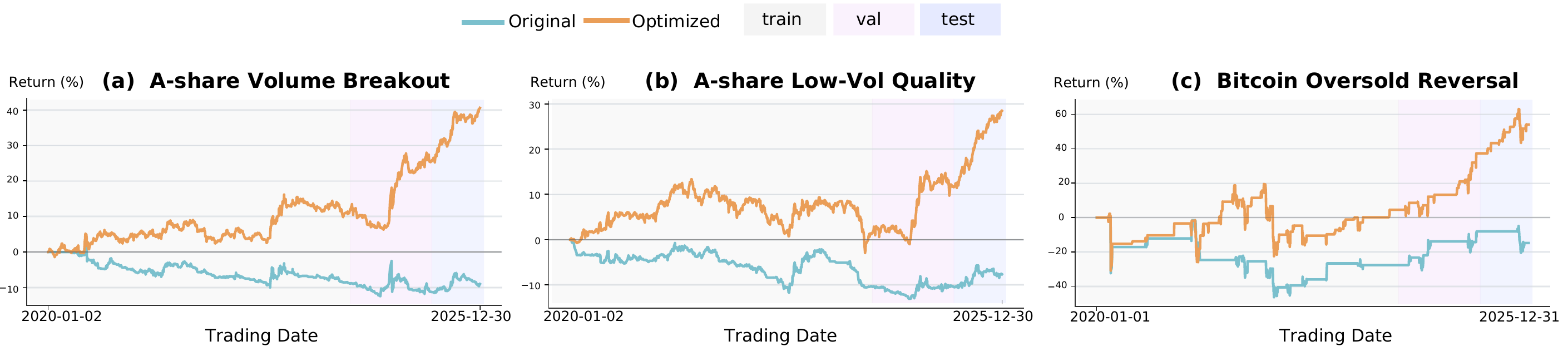}
\caption{Return curves of the original and optimized strategies on three representative tasks from 2020 to 2025: (a) A-share Volume Breakout, (b) A-share Low-Vol Quality, and (c) Bitcoin Oversold Reversal. Shaded regions denote the train, validation, and test periods.}
\label{fig:perf_before_after}
\end{figure*}

\begin{table}[!t]
\centering
\small
\resizebox{0.48\textwidth}{!}{

\begin{tabular}{@{}lrrr@{}}
\toprule
Volume Breakout                  & Sharpe $\Delta$ & Return $\Delta$ & MDD $\Delta$ \\ \midrule
\method{} Fully loaded                     & 0.982           & 17.95           & 1.81         \\ \midrule
\rowcolor{gray!10}[0pt][0pt]\multicolumn{4}{@{}l@{}}{\textit{Setting variations}} \\

Simple threshold                 & 0.053           & 3.39            & 3.34         \\
Parameter only                   & 0.274           & 6.59            & -0.21        \\
Random mutation only             & 0.645           & 13.05           & 2.34         \\
Bayesian Opt. parameter-only      & 0.739           & 11.11           & 1.14         \\ \midrule
\rowcolor{gray!10}[0pt][0pt]\multicolumn{4}{@{}l@{}}{\textit{Pipeline variations}} \\
w/o promotion engine             & 0.128           & 4.32            & 2.72      \\
w/o rule diagnosis               & 0.693           & 9.99            & -2.76        \\
w/o memory                       & 0.782           & 15.95           & -2.19        \\ \bottomrule
\end{tabular}}
\caption{Ablations on the A-share volume-breakout task. Returns and MDD are annualized percentage values.}
\label{tab:ablation}
\end{table}

\begin{table*}[!t]
\centering
\small
\resizebox{\textwidth}{!}{

\begin{tabular}{@{}clccccccccc@{}}
\toprule
\multirow{2}{*}{Protocol}                       & \multicolumn{1}{c}{\multirow{2}{*}{Strategy}} & \multicolumn{3}{c}{Sharpe $\uparrow$} & \multicolumn{3}{c}{Return (\%) $\uparrow$} & \multicolumn{3}{c}{MDD (\%) $\downarrow$} \\ \cmidrule(l){3-11} 
                                                & \multicolumn{1}{c}{}                          & Base       & Opt.       & $\Delta$    & Base        & Opt.        & $\Delta$       & Base        & Opt.        & $\Delta$      \\ \midrule
\rowcolor{gray!10}[0pt][0pt]\multicolumn{11}{@{}l@{}}{\textit{Robustness testing}} \\ \midrule
\multirow{4}{*}{2x fee/slip}                    & Bollinger Reversion                           & 0.647      & 1.025      & 0.378       & 8.71        & 11.40       & 2.69           & 2.62        & 2.25        & -0.37         \\
                                                & MA-RSI Trend                                  & -0.855     & -0.202     & 0.653       & -5.33       & 7.13        & 12.46          & 10.22       & 7.51        & -2.71         \\
                                                & Low-Vol Quality                               & -0.633     & 0.504      & 1.136       & -2.79       & 13.98       & 16.77          & 5.95        & 5.53        & -0.42         \\
                                                & Volume Breakout                               & -0.384     & 0.133      & 0.517       & -1.07       & 9.04        & 10.11          & 1.19        & 3.10        & 1.91          \\ \midrule
\multirow{4}{*}{Split shift}                    & Bollinger Reversion                           & 0.596      & 0.848      & 0.252       & 5.92        & 8.54        & 2.62           & 3.72        & 2.77        & -0.95         \\
                                                & EMA-RSI Trend                                 & -0.391     & 0.263      & 0.653       & -4.77       & 7.95        & 12.72          & 18.24       & 12.74       & -5.50         \\
                                                & Low-Vol Quality                               & -0.100     & 0.359      & 0.459       & 6.05        & 12.32       & 6.27           & 10.99       & 7.98        & -3.01         \\
                                                & Volume Breakout                               & -0.048     & -0.001     & 0.047       & 4.94        & 4.86        & -0.08          & 5.95        & 9.25        & 3.30          \\ \midrule
\rowcolor{gray!10}[0pt][0pt]\multicolumn{11}{@{}l@{}}{\textit{Walk-forward}} \\ \midrule
\multirow{4}{*}{2020--2023}                     & Bollinger Reversion                           & 0.099      & 0.355      & 0.256       & 1.09        & 5.51        & 4.42           & 1.00        & 1.44        & 0.44          \\
                                                & EMA-RSI Trend                                 & -0.408     & 0.440      & 0.848       & 4.26        & 21.03       & 16.77          & 5.38        & 4.76        & -0.62         \\
                                                & Low-Vol Quality                               & -0.156     & 0.306      & 0.462       & 8.00        & 11.13       & 3.13           & 4.02        & 4.19        & 0.17          \\
                                                & Volume Breakout                               & 0.159      & 0.196      & 0.038       & 1.15        & 1.43        & 0.28           & 0.00        & 0.03        & 0.03          \\ \midrule
\multirow{4}{*}{2021--latest}                   & Bollinger Reversion                           & 0.334      & 0.602      & 0.269       & 6.14        & 7.39        & 1.25           & 2.76        & 3.22        & 0.46          \\
                                                & EMA-RSI Trend                                 & -1.014     & -0.063     & 0.951       & -8.81       & 11.30       & 20.11          & 10.27       & 7.87        & -2.40         \\
                                                & Low-Vol Quality                               & -0.563     & 0.136      & 0.699       & -2.19       & 16.29       & 18.48          & 3.18        & 5.31        & 2.13          \\
                                                & Volume Breakout                               & 0.113      & 0.486      & 0.373       & 1.15        & 15.34       & 14.19          & 0.20        & 2.75        & 2.55          \\ \bottomrule
\end{tabular}
}
\caption{Robustness and rolling walk-forward portfolio results on the controlled slice. Returns and MDD are annualized percentage values.}
\label{tab:robustness}
\end{table*}

\subsection{Main A-Share Results}
Table~\ref{tab:main_astock_btc} reports the comparative results of the four quantitative strategies on the test set before and after optimization.We mainly report the performance of the strategy on the test set before and after optimization, because the test set is an important indicator for evaluating whether the optimized strategy has generalization ability. Figure~\ref{fig:perf_before_after}(a, b) present the comparison of the main quantitative strategies across all time periods. The remaining figures can be found in the appendix. All four strategy-family portfolios improve Sharpe after optimization, with Sharpe uplifts from \(-0.298\) to \(0.538\) and an average uplift of 0.829. Annual return also improves for all four strategies, with an average return uplift of 13.38 percentage points. Drawdown behavior is more divergent: EMA-RSI and low-volatility quality reduce MDD, while Bollinger reversion and volume breakout trade higher return for larger drawdown. 
The overall result shows that \method{} improves risk-adjusted performance broadly, but does not trivially optimize by only suppressing risk.

\subsection{Main Bitcoin Results}
Table~\ref{tab:main_astock_btc} and Figure~\ref{fig:perf_before_after}(c) report the Bitcoin results. All three original Bitcoin strategies start with negative test Sharpe. Optimization improves all three: MACD-RSI-Bollinger improves by 1.383 Sharpe points and becomes positive-Sharpe, oversold reversal improves by 1.516 and also becomes positive-Sharpe, while ATR trend breakout remains negative but improves by 0.486. Unlike the A-share portfolios, all three Bitcoin strategies reduce MDD, by 2.05 to 9.68 percentage points, suggesting that the verified edits improve both payoff quality and downside control in this market.

\subsection{Ablation Study}
Table~\ref{tab:ablation} isolates the main mechanisms on the fixed 30-stock A-share volume-breakout task. The full system achieves Sharpe uplift 0.982. Removing the promotion engine reduces Sharpe uplift to 0.128, while simple-threshold promotion reaches only 0.053. These results demonstrate that adaptive promotion serves as a core enabler. Simply expanding candidate pools cannot yield performance gains, since the non-promotion configuration involves more edits on average yet achieves markedly inferior results.

Candidate generation also plays an important role, although its effect is more nuanced. Restricting the search to parameter-only edits improves the Sharpe ratio by \(0.274\), indicating that local tuning alone cannot explain the full performance gain. The random-mutation baseline achieves an average improvement of \(0.645\). While this baseline remains competitive, it still underperforms the full system and is more prone to severe strategy drift and weak interpretability. \add{The Bayesian Optimization parameter-only baseline reaches an uplift of \(0.739\): even principled parameter search within the original strategy logic under the same verifier protocol cannot close the gap to the full system, confirming that the larger gain stems from verifier-constrained structural rewriting.} Removing the rule-diagnosis module reduces the improvement to \(0.693\), and removing the shared-memory mechanism further changes the improvement to \(0.782\). These results confirm that both diagnosis-guided search and memory-based experience reuse contribute materially to the overall system.

\subsection{Robustness and Walk-Forward}
Table~\ref{tab:robustness} reports stress and rolling evaluations. Under doubled fee/slippage assumptions, mean Sharpe uplift remains 0.700 , indicating that the gains are not purely a friction artifact. Under an alternative chronological split, the uplift falls to 0.368 but remains positive improvement. Rolling walk-forward evaluation also remains positive: the 2020--2023 window yields 0.370 uplift and the later-start window yields 0.545 uplift. The reduced magnitude under split shifts and rolling windows is expected; the important result is that performance weakens without collapsing.

\section{Discussion}
\stitle{Capability across strategies and markets.}
The results suggest that \method{} can optimize multiple strategy families rather than a single hand-picked rule. In the A-share market, the method improves diverse equity strategies after aggregating stock-level optimizations into portfolio-level results. In the cryptocurrency market, it also improves Bitcoin strategies with different volatility and trend characteristics, indicating broader applicability across market settings.

\stitle{Improvement is not just random edits.}
The ablation study indicates that the main benefit does not come from blindly perturbing strategy parameters or rules. Random mutation can occasionally find useful changes, but it lacks the failure-mode diagnosis, mechanism-aware hypothesis generation, and rejection feedback needed for reliable optimization. In contrast, LLM-guided strategy optimization contributes by turning backtest evidence into targeted edit hypotheses, explaining why an edit should help, and embedding those hypotheses in an evidence-aware promotion protocol.

\stitle{Verification is central to the contribution.}
The gains should be interpreted as risk-adjusted improvements under explicit evidence gates, not as universal drawdown reduction. Some A-share strategies accept higher drawdown in exchange for larger return and Sharpe, while the robustness tests show that the effect weakens under stricter settings without disappearing. This positions LLM-enabled strategy optimization as a verification problem as much as a generation problem: fluent edits are easy to produce, but trustworthy evidence requires chronological validation, provenance tracking, adaptive promotion, and visible rejection logs.

\add{\stitle{Methodological boundaries.}
\method{} claims relative improvement under a controlled historical backtesting protocol, not live-trading readiness or absolute guarantees under novel market regimes. Daily-bar backtests omit liquidity, borrow, and market-impact frictions, which bounds the absolute interpretation of the results while leaving the paired comparisons largely unaffected.}

\section{Conclusion and Future Work}
This paper studied automated quantitative strategy optimization as a verifier-guided program-evolution problem. Across A-share and Bitcoin strategies, \method{} improves risk-adjusted performance while keeping the optimization process auditable through explicit evidence gates and recorded edit decisions.

Future work can extend the evaluation to more realistic execution assumptions, larger investable universes, and stronger verifiers that combine statistical robustness tests with domain-specific financial constraints. These extensions would further clarify the boundary between useful automated strategy improvement and overfitting-driven search. \add{We also plan to integrate statistical overfitting tests such as the deflated Sharpe ratio into promotion, and to explore broader strategy discovery from typed hypothesis templates under the same verifier-guided constraint layer.}

\section*{Limitations}

This study is a research prototype, not an investment recommendation.
The backtests use local daily data and simplified execution assumptions.
Although we stress fees and slippage, real trading introduces liquidity, market impact, survivorship, corporate action, borrow, and operational constraints.
\add{To mitigate survivorship bias, we re-validated the A-share evaluation on a Wind point-in-time, delisting-inclusive universe (100 randomly sampled stocks for two strategy families); the improvement over the human-written baselines persists.}
The LLM comparison is also limited by local model/API configuration and budget.
\add{Since original and optimized strategies share the same data, splits, and execution assumptions, simplified daily-bar execution affects absolute interpretation more than the paired comparisons; richer execution modeling remains future work.}
These limitations do not invalidate the system contribution, but they bound the financial interpretation of the results.

\section*{Ethical Considerations}

This work studies automated strategy optimization for quantitative trading.
The methods are evaluated exclusively on historical backtests and do not constitute financial advice or a live trading system.
No personally identifiable information, proprietary data, or confidential market data were used.
All equity and cryptocurrency price data used in experiments are publicly available daily OHLCV records.
Any third-party models, APIs, datasets, or market data sources used by the artifact remain subject to their original licenses and terms of use.

The paper does not claim that the presented system is safe for deployment in real financial markets.
As discussed in the Limitations section, the evaluation relies on simplified execution assumptions that do not account for liquidity, market impact, or survivorship bias.
Practitioners who build on this work bear full responsibility for compliance with applicable financial regulations and risk management requirements.


\bibliography{references}

\appendix

\section{Prompt Templates}
\label{app:prompt}

This appendix lists the prompt templates used by \method{} for LLM-enabled candidate generation and diagnosis.
All prompts were used verbatim in the frontier LLM experiments reported in Section~5.4.

\subsection{Diagnosis Prompt}

The diagnosis prompt receives the evidence pack (returns, Sharpe, drawdown, trade counts, recent rejection reasons) and produces a structured bottleneck report.

\begin{tcolorbox}[
colback=gray!5, colframe=gray!50,
title=Promotion Explanation Prompt Template,
nobeforeafter,
breakable,
left=1mm,
right=1mm,
top=1mm,
bottom=1mm,
fontupper=\scriptsize,
fonttitle=\small\bfseries,
halign=left
]
\begin{verbatim}
You are a quantitative strategy analyst.
Given the following backtest evidence pack for 
strategy {strategy_name}:

  Train Sharpe  : {train_sharpe:.3f}
  Val Sharpe    : {val_sharpe:.3f}
  Test Sharpe   : {test_sharpe:.3f}
  Max Drawdown  : {max_dd:.2%}
  Trade Count   : {n_trades}
  Recent rejections: {rejection_summary}

Identify the top-2 bottlenecks from:
  [sparse_entry, delayed_exit, excessive_drawdown,
   low_trend_filter, overfit_validation, 
   parameter_sensitivity]

Return a JSON object:
{
  "bottlenecks": ["<bottleneck_1>", "<bottleneck_2>"],
  "rationale": "<one sentence per bottleneck>"
}
\end{verbatim}
\end{tcolorbox}

\subsection{LLM Candidate Generation Prompt}

The candidate generation prompt receives the current genome and the diagnosed bottlenecks, and proposes a single edited genome as a JSON diff.

\begin{tcolorbox}[
colback=gray!5, colframe=gray!50,
title=Promotion Explanation Prompt Template,
nobeforeafter,
breakable,
left=1mm,
right=1mm,
top=1mm,
bottom=1mm,
fontupper=\scriptsize,
fonttitle=\small\bfseries,
halign=left
]
\begin{verbatim}
You are a quantitative strategy engineer.
Current genome (JSON):
{genome_json}

Diagnosed bottlenecks: {bottlenecks}

Propose ONE targeted edit to address the primary bottleneck.
Constraints:
  - Do NOT change the core hypothesis of the strategy.
  - Only modify parameters or add/remove a single indicator.
  - Keep trade count above 20 per year.

Return ONLY a JSON patch in the format:
{
  "layer": "<signal|risk|position|exit>",
  "field": "<field_name>",
  "old_value": <old>,
  "new_value": <new>,
  "rationale": "<one sentence>"
}
\end{verbatim}
\end{tcolorbox}

\subsection{Promotion Explanation Prompt}

When a candidate is promoted, \method{} optionally queries the LLM for a human-readable explanation of the change, which is stored in the audit trail.

\begin{tcolorbox}[
colback=gray!5, colframe=gray!50,
title=Promotion Explanation Prompt Template,
nobeforeafter,
breakable,
left=1mm,
right=1mm,
top=1mm,
bottom=1mm,
fontupper=\scriptsize,
fonttitle=\small\bfseries,
halign=left
]
\begin{verbatim}
The following genome edit was accepted by the verifier:
  Layer    : {layer}
  Field    : {field}
  Old value: {old_value}
  New value: {new_value}
  Val Sharpe before: {val_before:.3f}
  Val Sharpe after : {val_after:.3f}

Summarize in one sentence why this edit is likely beneficial,
referencing the strategy hypothesis.
\end{verbatim}
\end{tcolorbox}

\section{Additional Experimental Results}
\label{app:additional_exp}

This appendix provides supplementary experimental results that support the claims in the main paper but were omitted from the main text due to space constraints.

\subsection{Main A-Share and Bitcoin Results}
\label{app:main_result}
\begin{figure*}[t]
\centering
\includegraphics[width=0.8\linewidth]{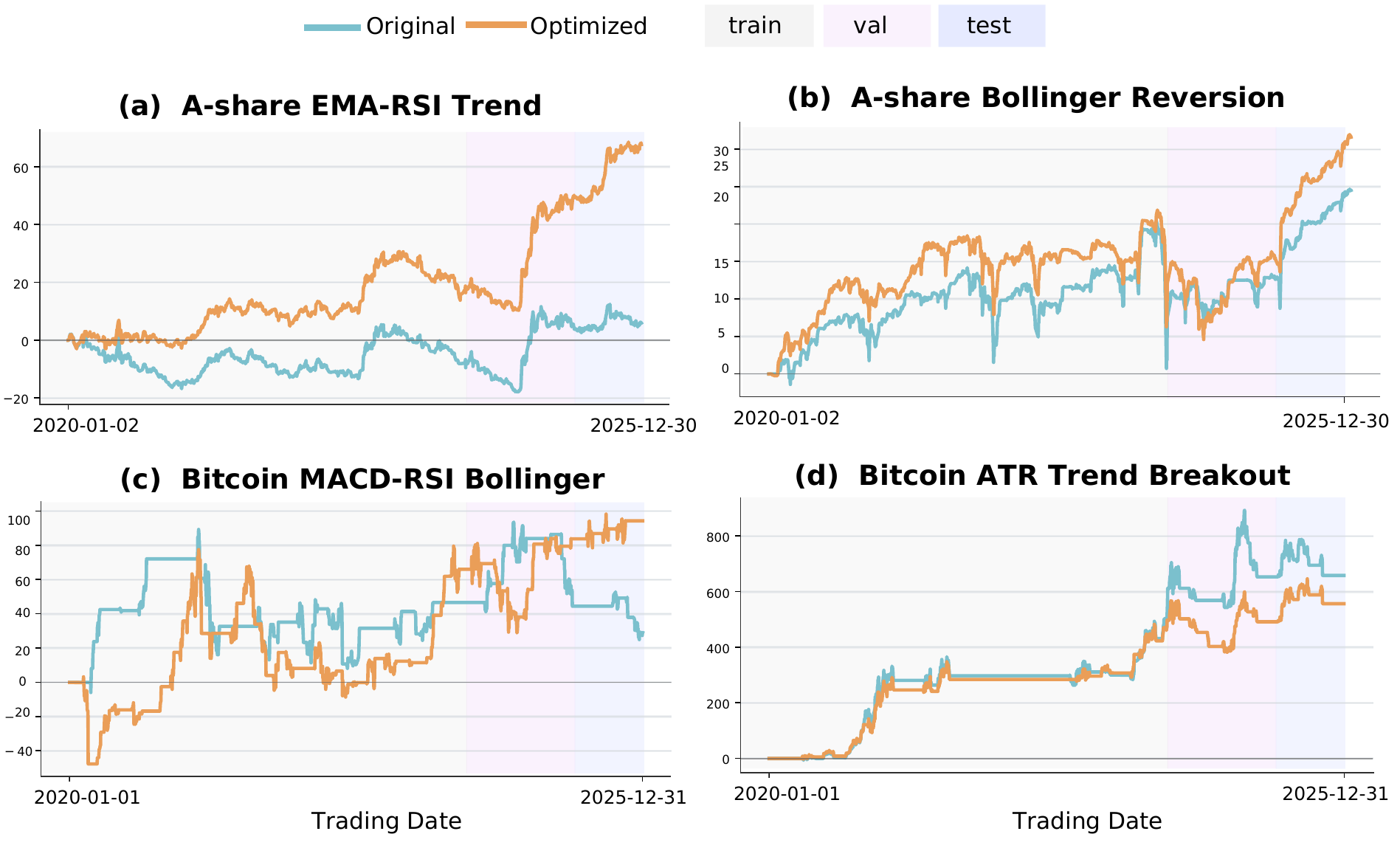}
\caption{Return curves of the original and optimized strategies on four representative tasks from 2020 to 2025: (a) A-share EMA-RSI Trend, (b) A-share Bollinger Reversion, (c) Bitcoin MACD-RSI Bollinger, and (d) Bitcoin ATR Trend Breakout. Shaded regions denote the train, validation, and test periods.}
\label{fig:perf_before_after_other}
\vspace{-1em}
\end{figure*}

Figure~\ref{fig:perf_before_after_other} shows that the optimized strategies generally achieve stronger return trajectories than their original counterparts across both A-share and Bitcoin tasks. 
For the A-share EMA-RSI Trend and Bollinger Reversion strategies, optimization leads to higher cumulative returns throughout the validation and test periods, with especially clear gains in the later test window. 
For Bitcoin strategies, the effect is more divergent: the optimized MACD-RSI Bollinger strategy improves substantially and ends with a higher return, whereas the optimized ATR Trend Breakout strategy underperforms the original strategy in the test period despite remaining profitable.

\subsection{Per-Symbol Sharpe Uplift Distribution}
\begin{figure}[H]
  \centering
  \includegraphics[width=\columnwidth]{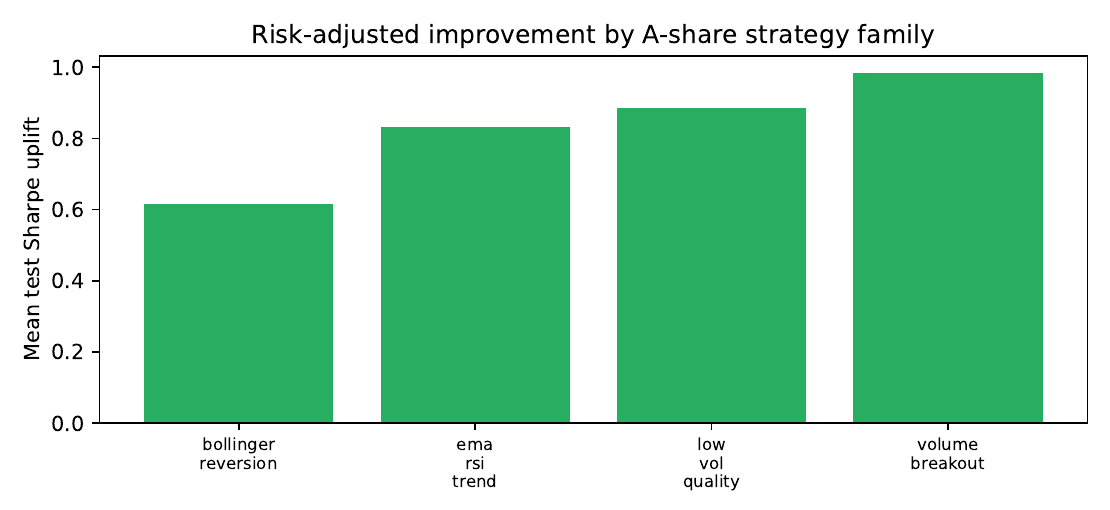}
  \caption{Distribution of per-symbol test Sharpe uplift (seed 42, all 120 A-share tasks).}
  \label{fig:app:sharpe_dist}
\end{figure}

Figure~\ref{fig:app:sharpe_dist} shows the full distribution of per-symbol test Sharpe uplift across all 120 A-share tasks (\ie task counted by the multiplication of 30 A-share stocks and 4 typical strategies) in the canonical seed-42 run.
The distribution is right-skewed, with a long tail of large improvements and a small mass of degradations, consistent with the 115/120 non-degradation rate reported in the main text.
The mean test Sharpe uplift is positive in every case, indicating that the gains are not driven by a single strategy type. 
The improvement is strongest for the Volume Breakout family, followed by Low-Vol Quality and EMA-RSI Trend, while Bollinger Reversion shows a smaller but still clear uplift. 
Overall, these results suggest that the verifier-guided optimization process can generalize across different A-share strategy families rather than only benefiting one specific signal design.

\FloatBarrier  

\subsection{Rejection Reasons}
\begin{figure}[H]
  \centering
  \includegraphics[width=\columnwidth]{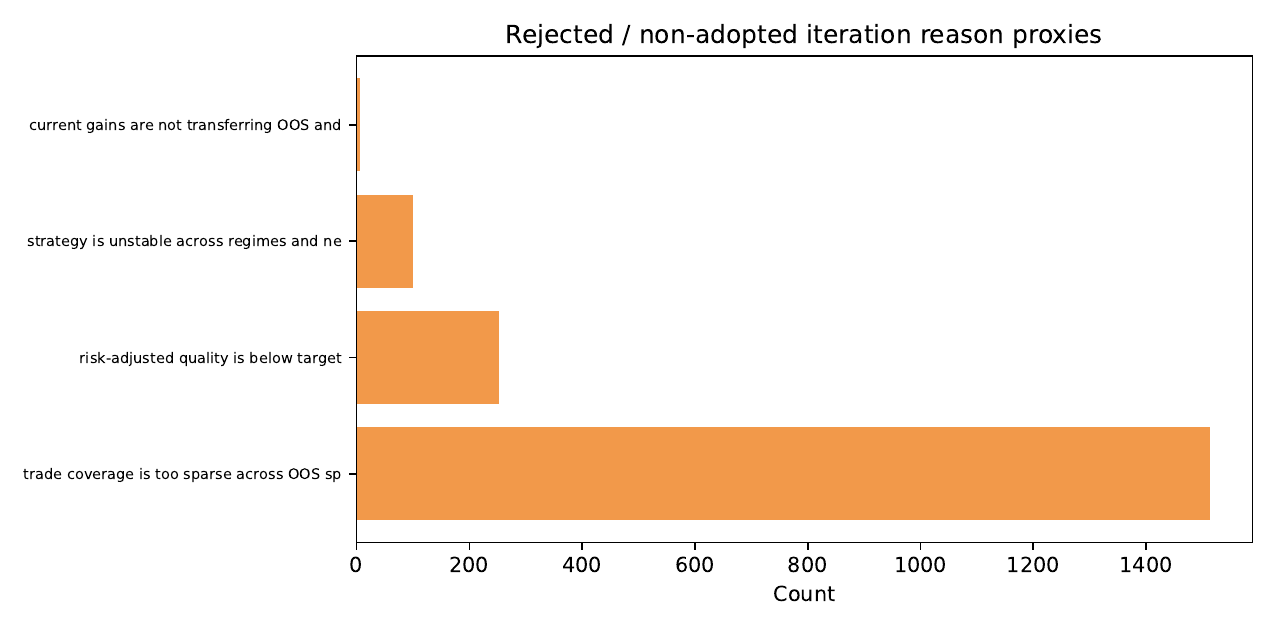}
  \caption{Distribution of rejection reasons for non-adopted candidate strategies. 
  Rejected candidates fail because: trade coverage is too sparse, below-target risk-adjusted quality, instability across regimes, or current gains that do not transfer out of sample.}
  \label{fig:app:rej}
\end{figure}

We further analyzed the rejection reasons across all experiments to characterize the optimizer's failure modes and the verifier's screening behavior.
The rejection analysis (see Figure~\ref{fig:app:rej}) reveals how the verifier filters unsafe or unreliable candidate strategies. 
The most frequent rejection reason is sparse trade coverage across OOS splits, indicating that many candidates do not generate enough out-of-sample trades to support reliable evaluation. 
A smaller number of candidates are rejected because their risk-adjusted quality falls below the target or because their performance is unstable across market regimes. 
Only a few candidates are rejected solely because current-window gains fail to transfer OOS.

\end{document}